\documentclass[11pt, a4paper, onecolumn, copyright]{main}

\usepackage[authoryear, round]{natbib}
\usepackage{xspace}
\usepackage{listings}
\usepackage{mathtools}
\usepackage{xurl}
\usepackage{fontawesome5}

\definecolor{darkgreen}{RGB}{50,100,0}
\definecolor{darkred}{RGB}{200,0,0}

\newcommand{\GEAR}{\textsc{GEAR}}
\titlespacing*{\paragraph}{0pt}{1ex plus 0.5pt minus 1pt}{1em}

\title{\GEAR{}\,\includegraphics[scale=0.12]{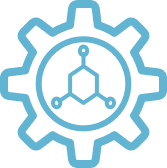}\,: Genetic AutoResearch for Agentic Code Evolution}

\runningtitle{\GEAR{}: Genetic AutoResearch}

\renewcommand{\today}{May 8, 2026}

\author[1, 2,*]{Ahmadreza Jeddi}
\author[1, 2,*]{Minh Ngoc Le}
\author[3]{Hakki C. Karaimer}
\author[3, 4]{Konstantinos G. Derpanis}
\author[1, 2]{Babak Taati}

\affil[1]{University of Toronto}
\affil[2]{Vector Institute}
\affil[3]{AI Center-Toronto, Samsung Electronics}
\affil[4]{York University}

\correspondingauthor{ajeddi@cs.toronto.edu}

 \definecolor{customred}{HTML}{ED028C}  

\hypersetup{
    colorlinks=true,
    urlcolor=customred  
}

\begin{abstract}
Autonomous research agents such as AutoResearch can now propose, run, and commit machine learning experiments without human supervision. However, their outer-loop search is largely \emph{single-incumbent hill climbing}: the agent edits one program, evaluates it, and retains the result only if it improves over the current best. We argue that this strategy prematurely discards valuable search signal, including complementary local optima, partially successful ideas, and accumulated insights from diverse research directions. We introduce \textbf{GEAR} (GEnetic AutoResearch), a drop-in search controller that replaces single-incumbent hill climbing with a population-based frontier search over research states. GEAR maintains a bounded population of elite nodes, selects parents using a composite score that balances estimated productivity, local novelty, and global coverage, and expands the frontier through mutation and crossover. Each node in the search graph preserves code changes, reflections, and performance statistics that inform future expansion decisions. We study three variants: (i) \textbf{GEAR-Prompt}, where the LLM agent manages population dynamics through natural-language instructions alone; (ii) \textbf{GEAR-Fixed}, which externalizes the genetic search policy into a fixed programmatic controller; and (iii) \textbf{GEAR-Evolve}, which treats the controller itself as a mutable artifact and explicitly decides at each iteration whether to run an experiment or modify the search policy. Under identical environments and compute budgets, all three GEAR variants outperform the AutoResearch baseline and achieve lower validation bits-per-byte. Beyond final performance, GEAR exhibits a distinct search dynamic: while the baseline quickly converges to a single local optimum, GEAR variants continue discovering improvements over longer horizons. These results suggest that equipping autonomous research agents with explicit population structure and mutable search policies can meaningfully extend their capacity for sustained progress.

\vspace{1em}

Project page: \href{https://genetic-autoresearch.github.io/}{\textcolor{customred}{https://genetic-autoresearch.github.io/}}
\end{abstract}

\makeatletter
\begin{document}

\begingroup
\renewcommand{\thefootnote}{}%
\renewcommand{\@makefnmark}{}%
\maketitle
\footnotetext{* Equal contribution.}
\makeatother
\endgroup

\section{Introduction}
\label{sec:intro}

\begin{figure}[t]
\centering
\includegraphics[width=\linewidth]{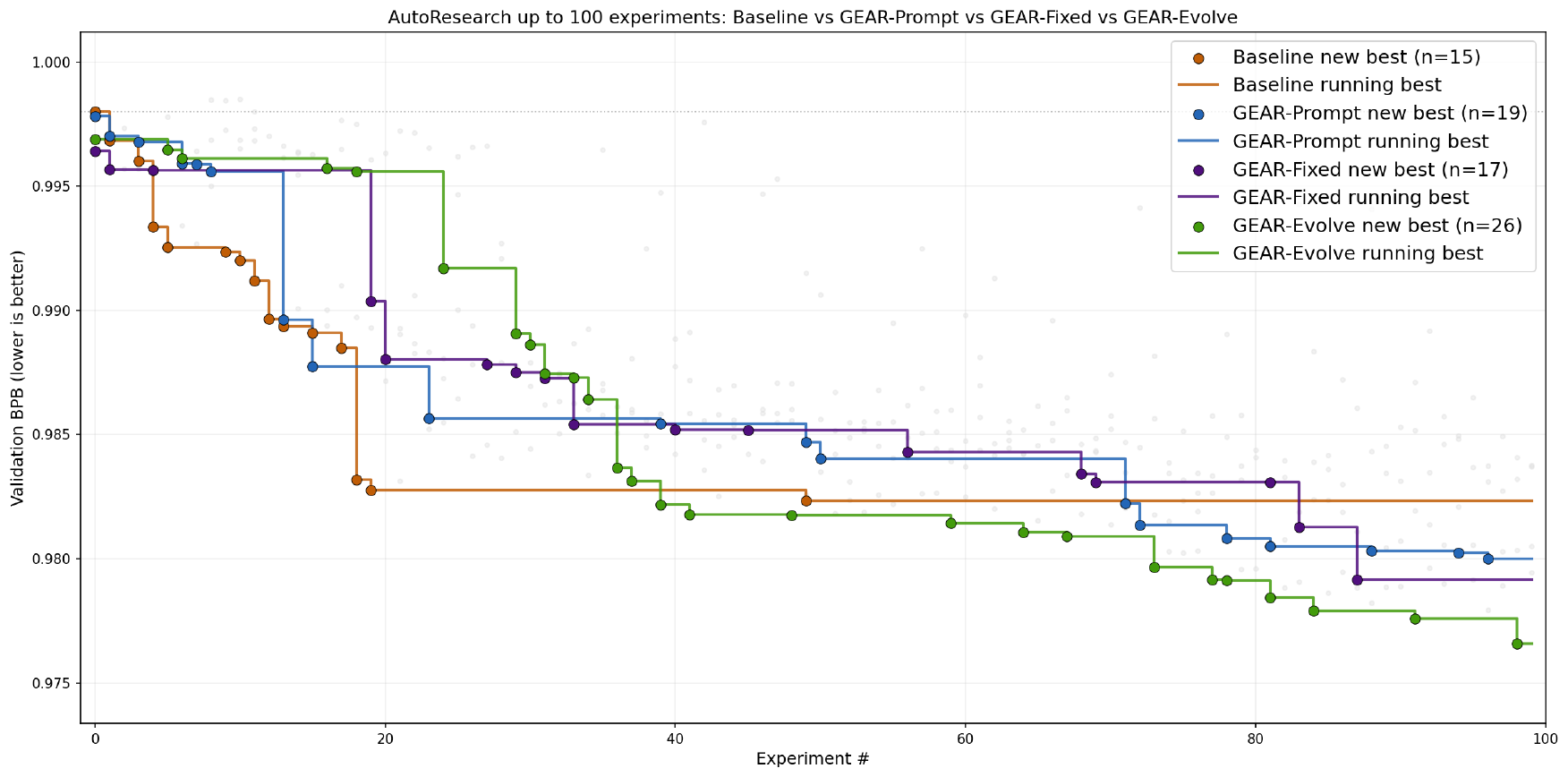}

\vspace{-0.3em}
\caption{Genetic AutoResearch (GEAR) variants all surpass the AutoResearch baseline's plateau after 100 experiments. Light grey dots are discarded experiments. Bold dots mark new global bests. \textsc{GEAR-Fixed} and \textsc{GEAR-Prompt} converge to similar quality, while \textsc{GEAR-Evolve} pulls ahead from roughly experiment 25 onward and eventually reaches the lowest bpb of the four.}
\label{fig:progress-compare}
\vspace{-1em}
\end{figure}

Automating machine learning research is shifting from simple code generation toward long-horizon experimental systems that can modify code, run training jobs, inspect outcomes, and decide what to try next \citep{lu2024aiscientist,yamada2025aiscientistv2,toledo2025ai,chen2026horizon,qu2026bilevel,borthwick2026robophd}. 
Among recent systems, AutoResearch~\citep{karpathy2026autoresearch} is a particularly simple and functional implementation of this idea, with a single LLM agent iteratively editing a training script, running under a fixed budget, and retaining changes that improve validation bits-per-byte (bpb) on a single GPU. Running Autoresearch without any human intervention for an extended period of time yields numerous additive improvements~\citep{karpathy2026autoresearch}.

The simplicity of this design, however, reveals a structural limitation.
The standard AutoResearch loop is effectively \emph{single-incumbent hill climbing}: there is one active artifact, one latest improvement, and a soft memory of prior attempts encoded only in logs or conversational context.
This is often sufficient for local progress, but it is a weak representation of research as a search process.
Real research typically spans multiple directions of investigation.
One direction may be strongest on raw performance, another may be leaner or more efficient, and a third might be temporarily uncompetitive yet contain an idea worth revisiting.
A single incumbent cannot represent these alternatives explicitly, and the ideas they embody are lost once discarded.

To this end, we propose \textbf{GEAR} (Genetic AutoResearch), a search controller designed as a drop-in replacement for the single-incumbent loop in AutoResearch-style systems.
GEAR preserves the original training environment and scalar objective but replaces keep-or-discard hill climbing with a search graph and a bounded frontier of best search nodes. Each node stores the code change and results, alongside its parentage, mutation type, and other descriptive metadata about how productive it has been as a source of future children, what it has tried and learned, and search statistics such as expansion count and improvement. The search graph is expanded through artifact mutation (proposing a code change from a single parent) and semantic crossover (synthesizing a child that combines complementary ideas from two parent nodes).

We study three variants of increasing structure: (i)~\emph{GEAR-Prompt}, where the LLM agent manages population dynamics through natural language instructions alone, (ii)~\emph{GEAR-Fixed}, which externalizes the search policy into a fixed programmatic controller that implements parent selection, promotion, and bookkeeping, and (iii)~\emph{GEAR-Evolve}, which additionally permits the agent to modify the controller code under logging and reproducibility constraints.

We evaluate all three variants on the same language modeling task used by AutoResearch and under identical compute budgets.
All GEAR variants consistently outperform the AutoResearch baseline.
More importantly, they exhibit a qualitatively different search dynamic: while the baseline quickly exhausts its useful search directions and settles into a local optimum, GEAR continues to discover improvements over longer horizons.
This sustained progress reflects the value of maintaining a frontier of diverse elite states rather than committing to a single incumbent.
By preserving partially successful ideas, revisiting complementary branches, and recombining useful changes, GEAR avoids the premature convergence that limits hill-climbing agents.

Our main contributions are as follows:
\begin{enumerate}[leftmargin=*,topsep=2pt,itemsep=1pt]
\item \textbf{GEAR}, a drop-in genetic search policy that replaces single-incumbent hill climbing with a bounded frontier of elite research states expanded through mutation and crossover, requiring no changes to the underlying training environment or evaluation harness (\S\ref{sec:gear-core}).

\item \textbf{A systematic dissection of genetic search mechanisms} in autonomous research agents, showing how mutation enables diverse local exploration while crossover composes complementary discoveries across branches. We instantiate this policy in three forms: natural-language prompting, a fixed programmatic controller, and a self-modifying controller that evolves its own search policy (\S\ref{sec:variants}).

\item \textbf{Empirical evidence} that population-based frontier search extends the horizon of autonomous discovery. While AutoResearch rapidly converges to a local optimum, GEAR variants continue making progress throughout the search budget, with the evolved controller producing the strongest sustained improvement (\S\ref{sec:experiments}).
\end{enumerate}

\section{Related Work}
\label{sec:related}
\subsection{Autonomous and Long-Horizon ML Research Agents}
\label{sec:related:agents}
Recent work on autonomous research has progressed from code assistance and paper drafting toward systems that execute substantial portions of the research loop end to end. Systems such as \emph{AI Scientist}~\cite{lu2024aiscientist, yamada2025aiscientistv2}, \emph{Agent Laboratory}~\cite{schmidgall2025agent}, \emph{AI-Researcher}~\cite{tang2025ai}, \emph{InternAgent}~\cite{feng2026internagent}, \emph{AI co-scientist}~\cite{gottweis2025towards}, and \emph{EvoScientist}~\cite{lyu2026evoscientist} study increasingly complete pipelines spanning literature review, hypothesis generation, code writing, experimentation, analysis, and manuscript preparation.
A closely related thread focuses more specifically on \emph{machine learning engineering} and long-horizon search in code space. \emph{AIDE}~\cite{jiang2025aide} formulates MLE as tree search over candidate code solutions, while methods like \emph{MLE-STAR}~\cite{nam2025mle} and ML-Master~\cite{liu2025ml, zhu2026toward}  augment this with memory, retrieval and targeted component-wise refinement. AIRA~\cite{toledo2025ai} makes the search-policy view explicit by studying how operator design and search strategy interact on MLE-bench~\cite{chan2024mle}, including greedy, Monte Carlo Tree Search (MCTS), and evolutionary search.  Within this landscape, Karpathy's \emph{AutoResearch}~\cite{karpathy2026autoresearch} is a particularly minimal and influential reference point: a single agent iteratively edits a training script, runs a bounded experiment, and keeps only improving changes. Many extensions of AutoResearch such as \cite{chen2026toward, he2026autoresearch} are already building on top of it.

This line of work is supported by a rapidly growing evaluation ecosystem. \emph{MLE-bench}~\cite{chan2025mlebench}, \emph{MLE-Dojo}~\cite{qiang2025mle}, \emph{ScienceAgentBench}~\cite{chen2024scienceagentbench}, and \emph{AIRS-Bench}~\cite{lupidi2026airs} evaluate agents on realistic MLE and scientific-discovery tasks with executable feedback and long-horizon iteration. There are also software-engineering benchmarks and agent frameworks such as \emph{SWE-bench}~\cite{jimenez2023swe}, \emph{ProgramBench}~\cite{yang2026programbenchlanguagemodelsrebuild}, \emph{SWE-agent}~\cite{yang2024swe}, \emph{OpenHands}~\cite{wang2024openhands}, and \emph{Agentless}~\cite{xia2024agentless} as adjacent evidence that repository-scale autonomy benefits from explicit tooling, execution feedback, and iterative repair, even when the end task is not necessarily scientific discovery. In this work, however, we only consider the baseline AutoResearch and their language modeling setup. Our focus is mainly on finding how the genetic search can impact agentic research in this minimal setup. 

\subsection{Evolutionary and Population-Based Search with LLMs}
\label{sec:related:evolution}
A second line of work studies how LLMs can participate in evolutionary or population-based optimization. At the prompt and inference level, \emph{OPRO}~\cite{yang2023large}, \emph{EvoPrompt}~\cite{guo2025evopromptconnectingllmsevolutionary}, \emph{Promptbreeder}~\cite{fernando2023promptbreederselfreferentialselfimprovementprompt}, \emph{EMO-Prompts}~\cite{baumann2024evolutionarymultiobjectiveoptimizationlarge}, \emph{GEPA}~\cite{agrawal2026gepa}, and \emph{Mind Evolution}~\cite{lee2025evolvingdeeperllmthinking} show that populations of candidate solutions, reflection, recombination, and selection can substantially improve prompts or reasoning trajectories.
At the program and algorithm level, recent methods increasingly combine evolutionary search with LLM-based generation and critique. \emph{ReEvo}~\cite{ye2024reevo} introduces reflective evolution for heuristic search, \emph{LLaMEA}~\cite{van2024llamea} evolves meta-heuristics with LLM-driven mutation and selection, and \emph{AlphaEvolve}~\cite{novikov2025alphaevolve} and \emph{ShinkaEvolve}~\cite{lange2025shinkaevolve} demonstrate strong performance in algorithmic code evolution with explicit populations and archives. Other recent works such as \emph{AVO}~\cite{chen2026avo}, \emph{Controlled Self-Evolution}~\cite{hu2026controlled}, \emph{TurboEvolve}~\cite{yang2026turboevolve}, and \emph{CodeEvolve}~\cite{assumpccao2025codeevolve} further explore LLM-driven mutation, crossover, memory, and multi-island search for code optimization.
These works establish the value of explicit populations, lineage, non-greedy exploration, and recombination. However, they are typically applied to prompt optimization, algorithm discovery, kernel optimization, or stand-alone code evolution rather than to long-horizon ML research loops with persistent experimental artifacts and a fixed training environment. GEAR sits at the intersection of these two literatures: it brings population-based evolutionary search into an AutoResearch-style experimental loop, while preserving the concrete, reproducible artifact structure needed for long-running machine learning research.

\section{Genetic AutoResearch}
\label{sec:gear}

\subsection{Setting}
\label{sec:setting}

GEAR uses the same experimental setting as AutoResearch~\citep{karpathy2026autoresearch}. The environment provides a single editable training file, and the goal is to train a GPT-2-style language model by minimizing validation bits-per-byte (bpb), which serves as the scalar performance objective.
Each experiment is subjected to a fixed training budget of five minutes on an NVIDIA H100 GPU, and the data pipeline, tokenizer, and evaluator harness are held constant across all runs. A Claude Opus~4.7~\citep{anthropic2026opus47} agent autonomously reads the repository, selects a parent, edits training code, launches the training job, records the outcome, and proceeds to the next iteration without human intervention.

\subsection{GEAR}
\label{sec:gear-core}

\begin{figure}[h]
\centering
\includegraphics[width=\linewidth]{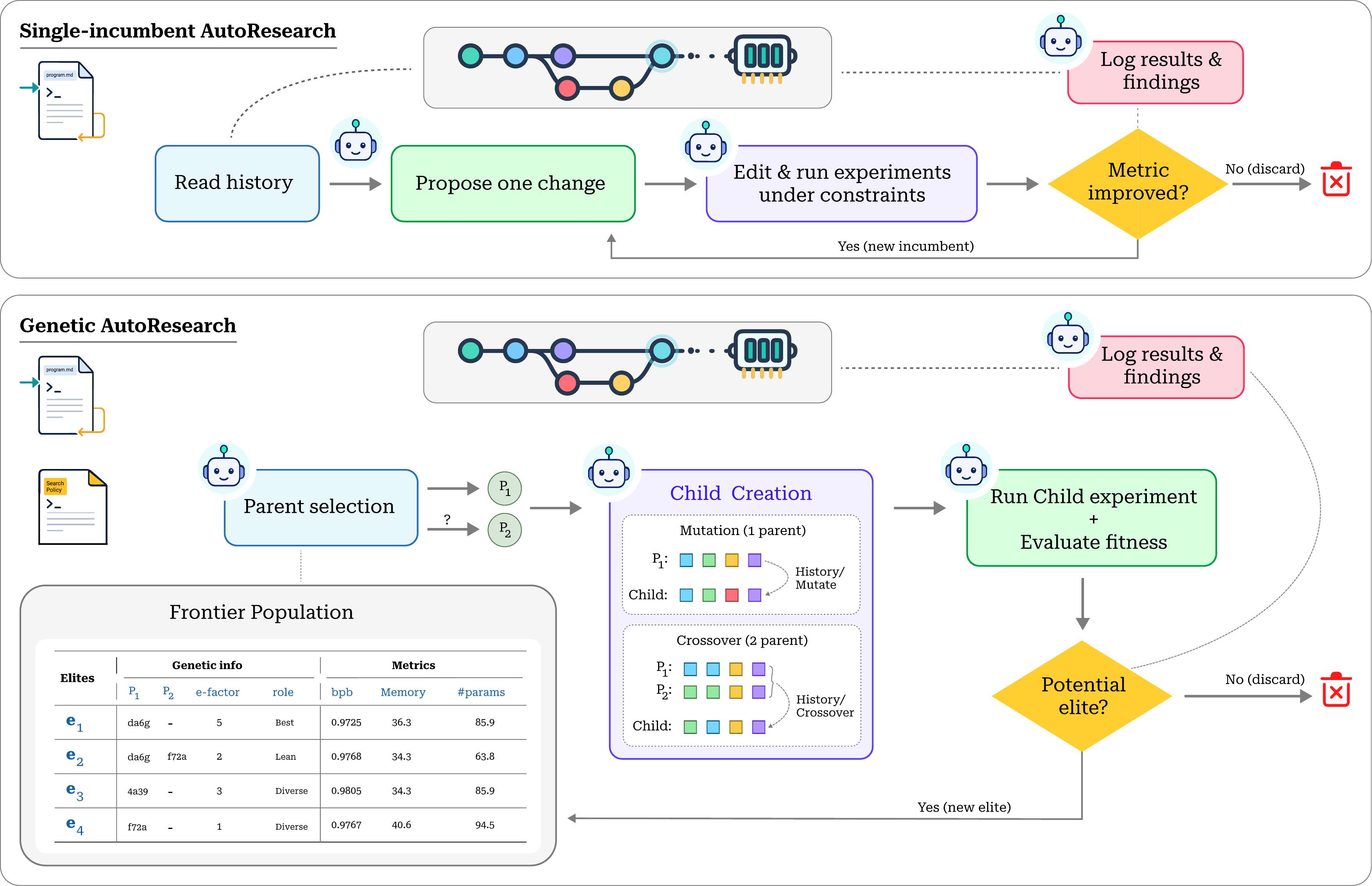}
\caption{In GEAR, the agent consults the frontier, selects a parent (or a pair of parents) trading off productivity, novelty, and coverage, spawns a child by mutation or crossover, runs the fixed training job, and either promotes the child into the frontier or discards it.}
\label{fig:gear-overview}
\end{figure}

GEAR replaces the single-incumbent hill climbing of AutoResearch with a population-based frontier search over research states.
The search maintains a frontier of nodes, each of which is a candidate solution. A node stores an edit of training code alongside its measured metrics, a description of what it tried, and statistics about how productive it has been as a parent.
Figure~\ref{fig:gear-overview} summarizes the resulting loop, in comparison to the baseline AutoResearch.

\paragraph{Search node.}
Let $\mathcal{F}_t = \{e_1, \dots, e_P\}$ denote the frontier with population $P$ at step $t$, and we refer to these nodes as elite nodes.
Each elite $e_i$ node represents one of the best research states discovered so far and contains a commit of the training code, its measured bpb, number of parameters, peak VRAM, a pointer to its parent(s), a short description of what it tried, node statistics such as the number of times $e$ has been used as a parent, the mean bpb improvement its children achieved over $e$, and the step at which it was last used.

\paragraph{Parent selection.}

At each step GEAR chooses a primary parent $p_1 \in \mathcal{F}_t$ subject to some constraints.
First, the same elite may not serve as $p_1$ in many consecutive steps while other elites exist, which prevents any single node from dominating the search for long stretches.
Second, over any short window of experiments, a minimum number of distinct elites must have served as $p_1$, so that all materially different nodes in the pool continue to produce children.
Third, the most recently used elite is mildly deprioritized, and a freshly promoted elite is mildly prioritized, so that new genetic material is exploited before it ages.
Periodically, to encourage diversity, GEAR additionally selects a secondary parent $p_2 \neq p_1$ that is maximally complementary to $p_1$, favoring elites with different roles and descriptions to $p_1$.
In \S\ref{sec:variants}, we present different variants of GEAR differing on whether they implement these genetic policies through natural language prompting or explicit functional code.


\paragraph{Spawning children: mutation and crossover.}

Given $p_1$, GEAR chooses between two operators: \emph{mutation} and \emph{crossover}.
For mutation, the agent proposes a change to the training code. Similar to the original AutoResearch loop, the agent may modify anything in the training script including model architecture, optimizer configuration, learning rate schedule, activation functions, hidden width, depth, batch size, or any other aspect.
For crossover, GEAR selects a second parent $p_2 \neq p_1$, and the agent, using $p_1$'s code as the base, transplants one coherent idea from $p_2$. $p_2$ is chosen to be materially different from $p_1$ to encourage diversity.
GEAR biases the operator choice so that crossovers appear regularly whenever at least two elites exist, and are prioritized immediately after a new elite is added so that fresh genetic material is recombined with the rest of the pool rather than left to drift on its own.

\paragraph{Roles.}
Each frontier slot is assigned one of three roles: \textsc{best} (lowest bpb), \textsc{lean} (lowest memory), or \textsc{diverse} (materially different description from all other elites).
The frontier is encouraged to contain at least one elite of each role at all times.
The \textsc{diverse} role's purpose is to keep a distinct line of investigation alive so that future crossovers have complementary material to recombine.
Without this role system, a frontier of elites can rapidly converge to minor variants of a single strong line, at which point the search is effectively back to single-incumbent hill climbing with a complicated bookkeeping system.

\paragraph{Promotion and discard.}
Once a child $c$ is spawned, its training code is run under the fixed five-minute budget and undergoes evaluation.
If the run completes successfully, GEAR decides whether $c$ should enter the frontier and, if so, which slot it should occupy.
Promotion follows a priority order.
If the frontier has an empty slot, the child fills it, which occurs during the first $P$ experiments as the population is built up.
If $c$ achieves a new global best bpb, it is promoted into the \textsc{best} slot.
If $c$ improves on the weakest elite, it takes that slot.
If $c$ is within noise of the best but uses less memory, it is promoted as the new \textsc{lean} elite.
Finally, if $c$ is only marginally worse than the weakest elite but its description is sufficiently distant from every current frontier member, it is promoted as a \textsc{diverse} elite to keep a materially different line of investigation alive for future crossovers.
Children that fail to clear any of these thresholds are discarded from the frontier but remain in the results log.
After every step, the agent writes a short reflection recording parent and child metrics, improvement deltas, the promotion decision, and a note on whether the idea merits revisiting under different conditions.

\begin{figure}[h]
\centering
\includegraphics[width=1\linewidth]{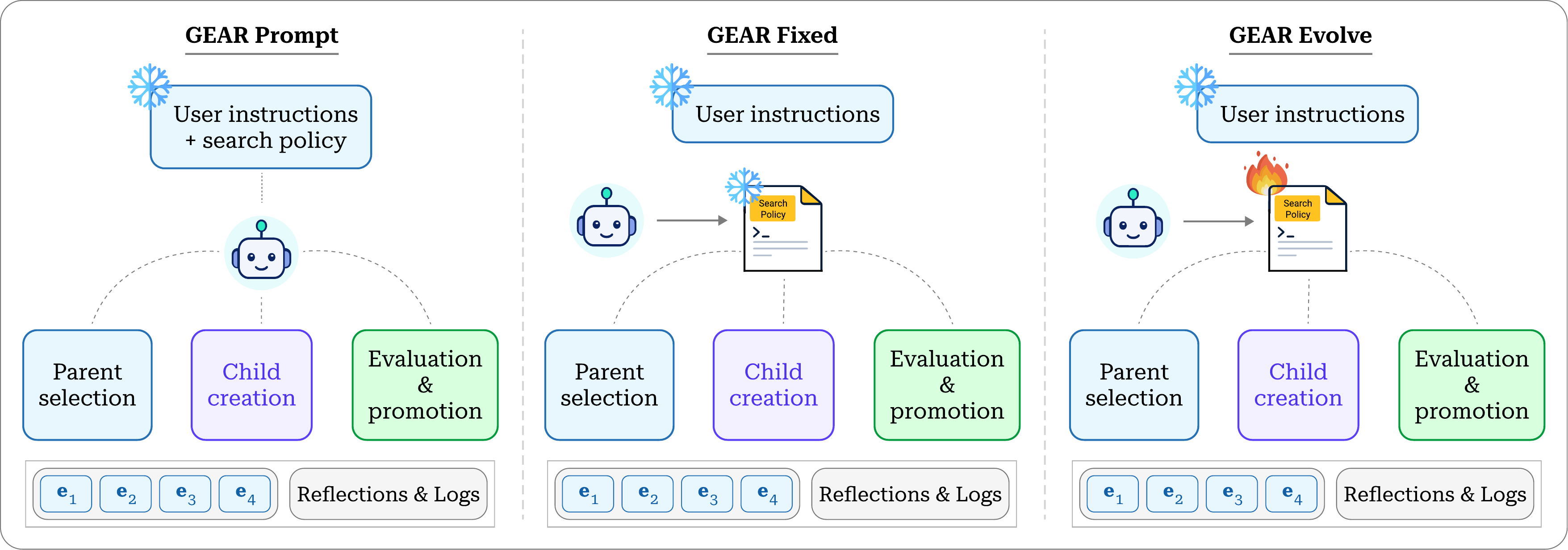}
\caption{\textbf{GEAR variants.} We study three implementations of the genetic search policy while keeping the underlying AutoResearch experimental setup fixed. In \textsc{GEAR-Prompt}, the policy is described in natural language and executed by the agent as part of its reasoning. In \textsc{GEAR-Fixed}, the policy is externalized into a deterministic controller that handles parent selection, operator scheduling, and promotion. In \textsc{GEAR-Evolve}, the controller itself becomes part of the search and can be modified by the agent over time.}
\label{fig:gear-variants}
\end{figure}

\subsection{Three Variants of GEAR}
\label{sec:variants}

The parent selection criteria and promotion rules, collectively referred to as \emph{genetic search policy}, can be specified to the agent in natural language or they can be externalized into a deterministic module. This external genetic search policy can be either an immutable file or it can be further evolved by the agent.
We study all three variants detailed below, holding the experiment setup identical to baseline AutoResearch. \autoref{fig:gear-variants} illustrates these variants, which share the same overall mechanics but differ in how the genetic search policy is implemented and executed.

\paragraph{GEAR-Prompt.}
The agent is given a natural language prompt that describes the population structure, the parent selection principles, the mutation and crossover operators, and the promotion logic, and is asked to execute the loop itself.
The full prompt is provided in Appendix~\ref{app:prompt-prompt}.
All bookkeeping are logged into persistent external files.
The selection criteria of \S\ref{sec:gear-core} appear as soft rules, such as ``do not use the same \texttt{parent1} in more than two consecutive experiments if another elite exists,'' ``when choosing a mutation parent, prefer the least recently used viable elite,'' ``at least two of any six consecutive non-crash experiments must be crossovers if two or more elites exist,'' and ``after creating a new elite, one of the next two non-crash experiments must be a crossover involving it.''
Promotion is similarly described in prose, with guidance to prioritize bpb first, memory second, and simplicity and diversity as tie-breakers.

\paragraph{GEAR-Fixed.}
This variant extracts the policy into a fixed externalized module that deterministically implements the decisions of what parents to choose, what child to spawn next, whether to promote a child, and how to update node statistics.
On each iteration, the agent requests parent(s), edits training code in the indicated parent's direction, runs the training job, and then records the outcome. 
This externalized search policy computes parent selection, operator choice, and promotion deterministically.
The policy that this controller implements has the following components:

\textit{Composite score.}
The controller selects the primary parent $p_1$ by scoring every elite $e \in \mathcal{F}_t$ and choosing the highest:

\begin{equation}
\mathrm{score}(e) \;=\; \underbrace{\bar{g}_e + \beta \sqrt{\tfrac{\log(N_t + 2)}{n_e + 1}}}_{\text{productivity}} \;+\; \lambda\, \mathrm{nov}(e) \;+\; \gamma\, \mathrm{cov}(e) \;+\; \rho(e).
\label{eq:score}
\end{equation}

The first term is a UCB-style productivity score. $\bar{g}_e$ denotes the mean bpb improvement that children of $e$ achieve relative to $e$ itself, $n_e$ is the number of times $e$ has served as a parent, and $N_t = \sum_{e' \in \mathcal{F}_t} n_{e'}$ is the total expansion count across the frontier at step~$t$.
The exploration bonus $\beta\sqrt{\log(N_t+2)/(n_e+1)}$ ensures that under-explored elites are not starved of children, following upper-confidence selection in bandit algorithms and tree  search~\citep{auer2002finite,kocsis2006bandit} and closely related to quality-diversity methods that formulate archive sampling as a bandit problem~\citep{sfikas2021monte}. The remaining terms add explicit novelty and
coverage pressures, following the broader quality-diversity and novelty-search literature~\citep{lehman2011abandoning,mouret2015illuminating}. The weights $\beta, \lambda, \gamma$ are fixed constants; $\lambda$ and $\gamma$ are lowered after an initial exploration phase so that, once the frontier has stabilized, productivity dominates. The complete list of scoring hyperparameters is given in Appendix~\ref{app:hyperparams}.

\textit{Novelty.}
The novelty term $\mathrm{nov}(e)$ computes the minimum Jaccard distance between $e$'s description and those of recently used parents.
Let $T(d)$ denote the set of lowercased alphanumeric tokens extracted from a description string~$d$.
Then
\begin{equation}
\mathrm{nov}(e) \;=\; \min_{e' \in \mathcal{R}_t \setminus \{e\}}\; 1 - \frac{|T(d_e) \cap T(d_{e'})|}{|T(d_e) \cup T(d_{e'})|},
\label{eq:nov}
\end{equation}
where $\mathcal{R}_t$ is the set of elites that served as primary parent in a short recency window.

\textit{Coverage.}
$\mathrm{cov}(e) = 1/\sqrt{k_e + 1}$, where $k_e$ is the number of frontier members that share the same role (\textsc{best}, \textsc{lean}, or \textsc{diverse}) as $e$ and target the same broad mutation category (e.g.\ optimizer, architecture, or regularization changes).
This term gives higher scores to elites whose role and category combination is rare in the current frontier, encouraging the search to explore directions that are currently underrepresented.

\textit{Recency adjustment.}
$\rho(e)$ is a small deterministic bonus that discourages consecutive reuse of the same parent and mildly prioritises freshly promoted elites.
Concretely, if $e$ was used as primary parent within the last one or two steps, $\rho(e) = -0.2$ or $-0.05$ respectively, and $0$ otherwise.

\textit{Secondary parent.}
When a crossover is scheduled, the second parent $p_2$ is chosen to be maximally complementary to the primary parent $p_1$.
Each candidate $e \in \mathcal{F}_t \setminus \{p_1\}$ is scored by
\begin{equation}
\mathrm{comp}(e;\, p_1) \;=\; J(d_{p_1},\, d_e) \;+\; \alpha\,\mathbf{1}[r_e \neq r_{p_1}] \;-\; \mu\, u(p_1, e),
\label{eq:comp}
\end{equation}
where $J(\cdot,\cdot)$ is the Jaccard distance between description token sets defined in Eq.~\ref{eq:nov}, $r_e$ denotes the role of elite $e$, and $u(p_1, e)$ counts the number of prior crossovers in which $p_1$ and $e$ were co-used as parents.
The indicator term with weight $\alpha{=}0.2$ rewards role mismatch, and the penalty with weight $\mu{=}0.1$ discourages repeating the same parent pair.
The candidate with the highest complementarity score is selected as $p_2$.
Crossover is forced whenever fewer than one crossover has occurred in the last four experiments and at least two elites exist, and is also forced immediately after a promotion involving a new elite.

\textit{Promotion.}
Promotion uses an improvement margin $\epsilon_{\text{imp}} = 1.5\times 10^{-4}$ bits-per-byte for ``strictly better,'' a tie margin $\epsilon_{\text{tie}} = 1.2\times 10^{-4}$ for ``within noise of the best,'' a memory margin of $0.5\,$GB for the \textsc{lean} replacement rule, and a Jaccard threshold of $0.65$ for the \textsc{diverse} replacement rule.
The sequence of checks matches the prose description in \S\ref{sec:gear-core}.

\paragraph{GEAR-Evolve.}
This variant treats the genetic search policy itself as part of the search.
At every iteration, the agent must choose between an \emph{experiment step} (running a child according to the current search policy) or a \emph{controller step} (editing the search policy to improve the search).
Both choices are recorded in decision logs, which contain the choice and a short reason.
To prevent the agent from defaulting to experiment-only behavior, the protocol requires that after five consecutive experiment steps the next decision line either be a controller step or contain an explicit justification that the search is healthy.
The full prompt is provided in Appendix~\ref{app:prompt-evolve}.
\section{Experiments}
\label{sec:experiments}

\subsection{Setup}
\label{sec:exp-setup}

We compare the three variants of GEAR with the baseline AutoResearch~\cite{karpathy2026autoresearch} on the same code editing task (\S\ref{sec:setting}). All four runs use the same starting codebase, the same fixed five-minute training budget on a single NVIDIA H100 GPU, and the same evaluation harness.
We run each variant for 100 experiment steps.
The primary metric is bpb on the evaluation set, where lower is better. We report the running best bpb over experiments. Peak VRAM and number of parameters are reported as secondary axes.

\subsection{Main Result}
\label{sec:exp-main}

Figure~\ref{fig:progress-compare} plots the running best bpb of each variant against the experiment count. Table~\ref{tab:final-best} summarizes the final best elite of each run. All three GEAR variants outperform the AutoResearch baseline, and \textsc{GEAR-Evolve} is best on every axis.

\begin{table}[h]
\caption{Final best elite per variant within the first $100$ experiments. ``First exp.\ to beat Baseline'' is the index of the first experiment whose bpb falls below the Baseline's final value of $0.98232$. \textsc{GEAR-Evolve} reaches that level $44$ experiments earlier than \textsc{GEAR-Fixed} and $32$ earlier than \textsc{GEAR-Prompt}.}
\label{tab:final-best}
\centering
\small
\begin{tabular}{lrrrr}
\toprule
Variant & bpb $\downarrow$ & VRAM (GB) & Params (M) & First exp.\ to beat Baseline \\
\midrule
Baseline (AutoResearch) & 0.98232 & 60.2 & 80.9 & --- \\
\textsc{GEAR-Prompt}    & 0.98001 & 63.6 & 71.3 & 72 \\
\textsc{GEAR-Fixed}     & 0.97914 & 66.2 & 85.9 & 84 \\
\textsc{GEAR-Evolve}    & \textbf{0.97658} & \textbf{33.5} & 85.9 & \textbf{40} \\
\bottomrule
\end{tabular}

\end{table}

\paragraph{All three GEAR variants outperform the AutoResearch baseline.}
The Baseline produces no further improvements after experiment~50 (bpb $=0.98232$), despite the agent trying a wide range of architectural and optimizer changes across the remaining 50 experiments.
\textsc{GEAR-Prompt} reaches $0.98001$, \textsc{GEAR-Fixed} reaches $0.97914$, and \textsc{GEAR-Evolve} reaches $0.97658$. Evidently, our genetic search policy gives the agent more useful structure than a single greedy incumbent. With a frontier of distinct anchors, the agent revisits older directions under new conditions and continues finding improvements long after the baseline has stopped finding any.
Furthermore, allowing the agent to revise its own search policy pays off in both quality and sample efficiency: \textsc{GEAR-Evolve} is the first variant to cross Baseline's plateau (experiment 40), \textsc{GEAR-Prompt} crosses it at experiment 72, and \textsc{GEAR-Fixed} only at experiment 84 (Table~\ref{tab:final-best}).

\paragraph{Memory and parameter count.}
\textsc{GEAR-Evolve} produces a substantially leaner final model than the other three variants ($33.5$\,GB peak VRAM vs.\ $60.2$--$66.2$\,GB). The genetic search policy actively rewards elites that use less memory through the \textsc{lean} role, which biases the search toward more optimized directions.
\textsc{GEAR-Evolve} discovers early that halving the batch size to $2^{17}$ both improves bpb and frees enough VRAM to scale model depth, a compound move that becomes the foundation of its subsequent trajectory.
\textsc{GEAR-Fixed} and \textsc{GEAR-Prompt} converge on higher-memory configurations ($66.2$ and $63.6$\,GB respectively), having committed to depth and width settings before discovering the reduction of batch size.

\subsection{Discussion}
\label{sec:exp-where}

\paragraph{GEAR sustains improvement throughout the budget.}
Table~\ref{tab:quarterly} breaks each run into four blocks of $25$ experiments and reports the improvement in best bpb per block. The Baseline concentrates $97\%$ of its total improvement in the first quarter and produces zero improvement across the entire second half of its budget.
All three GEAR variants continue improving in every quarter.
\textsc{GEAR-Evolve} produces $9.95$\,mbpb in its second quarter and sustains $2$--$3$\,mbpb of improvement in each subsequent quarter. 
Our frontier search policy provides the structural diversity needed to avoid the premature saturation that consumes half the Baseline's budget.

\begin{table}[h]
\caption{Improvement in running best bpb (mbpb, higher is better) per block of 25 experiments. All GEAR variants sustain improvement throughout the budget, unlike Baseline which plateaued early.}
\label{tab:quarterly}
\centering
\small
\begin{tabular}{lrrrr}
\toprule
& \multicolumn{4}{c}{Improvement per quarter (mbpb)} \\
\cmidrule(lr){2-5}
Variant & $1$--$25$ & $26$--$50$ & $51$--$75$ & $76$--$100$ \\
\midrule
Baseline & 15.24 & 0.45 & 0.00 & 0.00 \\
\textsc{GEAR-Prompt} & 12.18 & 0.94 & 2.68 & 1.35 \\
\textsc{GEAR-Fixed} & 8.38 & 2.87 & 2.08 & 3.95 \\
\textsc{GEAR-Evolve} & 5.21 & 9.95 & 2.08 & 3.08 \\
\bottomrule
\end{tabular}

\end{table}

\paragraph{Compositional gains via staged exploration.}
Several of \textsc{GEAR-Evolve}'s largest improvements come from multi-step compositional trajectories that are structurally difficult to achieve in single-incumbent hill climbing. The clearest example is the batch-size-to-depth chain. At experiment~2, the agent tries increasing depth from~8 to~9, which regresses because the default large batch size under-trains the bigger model. The depth-9 configuration is retained in the frontier as a non-best elite. Over experiments~17--35, a separate line of investigation discovers that halving the batch size and adjusting the warmdown ratio produces large gains at depth~8. At experiment~37, the agent revisits depth scaling with smaller batch size, and depth~9 succeeds (bpb $0.984$, a $2.7$\,mbpb improvement). Depth~10 follows at experiment~38 and, after re-tuning the learning rate, surpasses the Baseline's final value at experiment~40.

This trajectory requires the frontier to preserve multiple directions simultaneously.  A single-incumbent system can in principle revisit failed ideas through its conversational ``soft memory'' --- and the Baseline does retry depth-9 successfully at experiment~18, after accumulating warmdown improvements --- but the Baseline tries depth-10 twice (experiments~32 and~71) and fails both times. All three GEAR variants succeed at depth~$\geq$10: \textsc{GEAR-Prompt} at experiment~24, \textsc{GEAR-Fixed} at experiment~32, and \textsc{GEAR-Evolve} at experiment~38. The frontier's ability to maintain multiple anchors allows each variant to compose depth scaling with batch-size, learning-rate, and schedule discoveries that make the deeper model viable.


\paragraph{Crossover usage.}
The three GEAR variants invoke crossover at comparable rates, but the variants with an externalized controller produce substantially higher-quality crossovers. In \textsc{GEAR-Prompt}, $28$ of $29$ crossovers use the same parent pair (\texttt{elite/0}, \texttt{elite/1}), and only $14\%$ produce an elite. The descriptions confirm that most reduce to single-hyperparameter transplants from a fixed secondary parent. The deterministic controllers in \textsc{GEAR-Fixed} and \textsc{GEAR-Evolve} enforce complementarity scoring and penalties for reusing the same parents (Eq.~\ref{eq:comp}), producing $35$ and $33$ distinct parent pairs respectively, with elite promotion rates of $71\%$ and $72\%$.
The value of mechanized crossover is illustrated by \textsc{GEAR-Fixed}'s strongest single improvement: at experiment~89, a mutation discovers Muon $\beta_2{=}0.85$ (bpb $0.983$), and the controller's next crossover suggestion at experiment~90 transplants this idea onto the accumulated best base, producing a $2.1$\,mbpb jump that accounts for the variant's final best. Without the controller's complementary pairing, this combination would have required the agent to independently identify and execute the transplant.

\paragraph{Evolved genetic search policy.}
\textsc{GEAR-Evolve}'s six controller edits within the $100$ experiments share a common theme: each repairs a case where the crossover operator produces a degenerate or redundant suggestion. The agent observes a concrete failure in the controller's output, diagnoses the root cause in the code, and patches the specific function responsible. Table~\ref{tab:controller-edits} lists the edits in chronological order.

\begin{table}[h]
\caption{The six controller edits made by \textsc{GEAR-Evolve} during the $100$ experiments. For each edit, the agent observes a degenerate crossover suggestion, identifies the responsible code, and patches it.}
\label{tab:controller-edits}
\centering
\small
\renewcommand{\arraystretch}{1.2}
\begin{tabular}{clll}
\toprule
\# & Before & Agent's observation & Suggested Fix \\
\midrule
1 & exp-3 & Crossover suggested between baseline & Block crossover until $\geq 2$ \\
  &       & and its only child  & non-baseline elites exist \\
2 & exp-4 & Parent and its direct mutate-child & Skip candidates that are direct \\
  &       & paired for crossover & ancestors of the primary parent \\
3 & exp-6 & Known-bad elite repeatedly chosen & Prefer the current best as primary \\
  &       & as primary, producing duplicates & until it has been expanded $\geq 3$ times \\
4 & exp-20 & Direct-parent check misses multi- & Walk the full parent chain via \\
  &       & hop ancestry (grandparent chains) & breadth-first search over results log \\
5 & exp-22 & Same pair re-suggested despite & Block crossover when no untried \\
  &       & pair penalty (penalty too weak) & independent pair remains \\
6 & exp-37 & Ancestry traced only via primary & Extend ancestry walk to follow \\
  &       & parent, missing ideas inherited & secondary parent pointers as well \\
  &       & through prior crossovers & \\
\bottomrule
\end{tabular}

\end{table}

The edits follow an architectural progression. Edits~1--3 are simple fixes that include blocking crossover when too few viable elites exist and validating parent selection. Edits~4--6 are more structural: the agent introduces an ancestry-tracking routine that performs a breadth-first walk over the full parentage graph stored in the results log, initially following only primary-parent pointers and later extending to secondary parents. This routine, which was not present in the original controller, was built from scratch in response to repeated encounters with the same class of failure. By experiment~37, the controller's crossover suggestions consistently combine genuinely complementary material, eliminating wasted iterations on redundant combinations.

\section{Conclusion}
\label{sec:exp-discussion}



In this work, we introduced \textsc{GEAR}, a genetic search framework for autonomous research agents that replaces single-incumbent hill climbing with population-based frontier search. Across three variants, \textsc{GEAR} forms a natural progression from \emph{policy in prompt}, to \emph{policy in code}, to \emph{policy as a search target}. This progression shows that moving from prompted instructions to a fixed controller turns parent rotation, role enforcement, and crossover from optional behaviors into enforced mechanisms. Making the controller mutable adds a further capability, allowing the agent to revise its own search policy when the current invariants are insufficient. Together, these results suggest that autonomous research agents benefit from explicit mechanisms for maintaining, recombining, and evolving diverse lines of inquiry.


\end{document}